\documentclass[final]{cvpr}
 
\usepackage[utf8]{inputenc} % allow utf-8 input
\usepackage[T1]{fontenc}    % use 8-bit T1 fonts
\usepackage{url}            % simple URL typesetting
\usepackage{booktabs}       % professional-quality tables
\usepackage{amsfonts}       % blackboard math symbols
\usepackage{nicefrac}       % compact symbols for 1/2, etc.
\usepackage{microtype}      % microtypography
\usepackage{times}
\usepackage{epsfig}
\usepackage{graphicx}
\usepackage{amsmath}
\usepackage{amssymb}
\usepackage{multirow}
\usepackage{wrapfig}
\usepackage{subfigure}
\usepackage{ntheorem}

\usepackage{float}
\usepackage{algorithm}
\usepackage{algorithmic}
\usepackage{wrapfig}
\usepackage{ragged2e}
\usepackage{changepage}
\usepackage{color}
\usepackage[pagebackref=true,breaklinks=true,colorlinks,bookmarks=false]{hyperref}

\def\cvprPaperID{636} % *** Enter the CVPR Paper ID here
\def\confYear{CVPR 2021}

\graphicspath{{../}}

\begin{document}

\title{HourNAS: Extremely Fast Neural Architecture Search \\ Through an Hourglass Lens}

\author{Zhaohui Yang$^{1,2}$, Yunhe Wang$^{2}$, Xinghao Chen$^{2}$, Jianyuan Guo$^{2}$, \\
	Wei Zhang$^{2}$, Chao Xu$^{1}$, Chunjing Xu$^{2}$, Dacheng Tao$^{3}$, Chang Xu$^{3}$ \\
	\normalsize$^1$ Key Lab of Machine Perception (MOE), Dept. of Machine Intelligence, Peking University.
	\\
	\normalsize$^2$ Noah's Ark Lab, Huawei Technologies. \\
	\normalsize$^3$ School of Computer Science, Faculty of Engineering, University of Sydney.\\
	\small\texttt{\{zhaohuiyang,jyguo\}@pku.edu.cn; xuchao@cis.pku.edu.cn}\\
	\small\texttt{\{yunhe.wang,xinghao.chen,wz.zhang,xuchunjing\}@huawei.com}\\
	\small\texttt{\{dacheng.tao,c.xu\}@sydney.edu.au}
}

\maketitle

\begin{abstract}
Neural Architecture Search~(NAS) aims to automatically discover optimal architectures. In this paper, we propose an hourglass-inspired approach~(HourNAS) for extremely fast NAS. It is motivated by the fact that the effects of the architecture often proceed from the vital few blocks. Acting like the narrow neck of an hourglass, vital blocks in the guaranteed path from the input to the output of a deep neural network restrict the information flow and influence the network accuracy. The other blocks occupy the major volume of the network and determine the overall network complexity, corresponding to the bulbs of an hourglass. To achieve an extremely fast NAS while preserving the high accuracy, we propose to identify the vital blocks and make them the priority in the architecture search. The search space of those non-vital blocks is further shrunk to only cover the candidates that are affordable under the computational resource constraints. Experimental results on ImageNet show that only using 3 hours~(0.1 days) with one GPU, our HourNAS can search an architecture that achieves a 77.0\% Top-1 accuracy, which outperforms the state-of-the-art methods.
\end{abstract}

\section{Introduction}

In the past decade, progress in deep neural networks has resulted in the advancements in various computer vision tasks, such as image classification~\cite{alexnet, vgg, multicolumndnn, stochasticdepth}, object detection~\cite{frcnn, ssd}, and segmentation~\cite{maskrcnn}. The big success of deep neural networks is mainly contributed to the well-designed cells and sophisticated architectures. For example, VGGNet~\cite{vgg} suggested the use of smaller convolutional filters and stacked a series of convolution layers for achieving higher performance, ResNet~\cite{resnet} introduced the residual blocks to benefit the training of deeper neural networks, and DenseNet~\cite{densenet} designed the densely connected blocks to stack features from different depths. Besides the efforts on the initial architecture design, extensive experiments are often required to determine the weights and hyperparameters of the deep neural network. 

To automatically and efficiently search for neural networks of desireable properties (\emph{e.g.}, model size and FLOPs) from a predefined search space, a number of Neural Architecture Search (NAS) algorithms~\cite{darts, yang2019evaluation, li2020random, yu2019evaluating, nasbench101, nasbench201} have been recently proposed. Wherein, Evolutionary Algorithm (EA) based methods~\cite{amoebanet} maintain a set of architectures and generate new architectures using genetic operations like mutation and crossover. Reinforcement Learning (RL) based methods~\cite{zoph_rl_iclr, nasnet} sample architectures from the search space and train the controllers accordingly. The differentiable based methods~\cite{darts, fbnet, snas, metaarchitecturesearch} optimize the shared weights and architecture parameters, which significantly reduces the demand for computation resources and makes the search process efficient.

\begin{figure*}[t]
	\centering
	\iffalse
	\subfigure[]{\label{fig_resnet}\includegraphics[height=5cm]{figs/fig_demo_multipathnetwork.png}}
	\subfigure[]{\label{fig_resnet_hourglass}\includegraphics[height=5cm]{figs/fig_hourglass.png}}
	\hspace{20pt}
	\subfigure[]{\label{fig_critical}\includegraphics[height=5cm]{figs/fig_critical.png}}
	\subfigure[]{\label{fig_minimal_supernet}\includegraphics[height=5cm]{figs/fig_demo_criticallayersearch.png}}
	\subfigure[]{\label{fig_non_critical_supernet}\includegraphics[height=5cm]{figs/fig_demo_noncriticallayersearch.png}}
	\subfigure[]{\label{fig_searched}\includegraphics[height=5cm]{figs/fig_demo_searched.png}}
	\subfigure{\label{fig_search_space}\includegraphics[width=0.7\linewidth]{figs/fig_demo_notations.png}}
	\fi
	\subfigure{\label{fig_search_space}\includegraphics[width=0.9\linewidth]{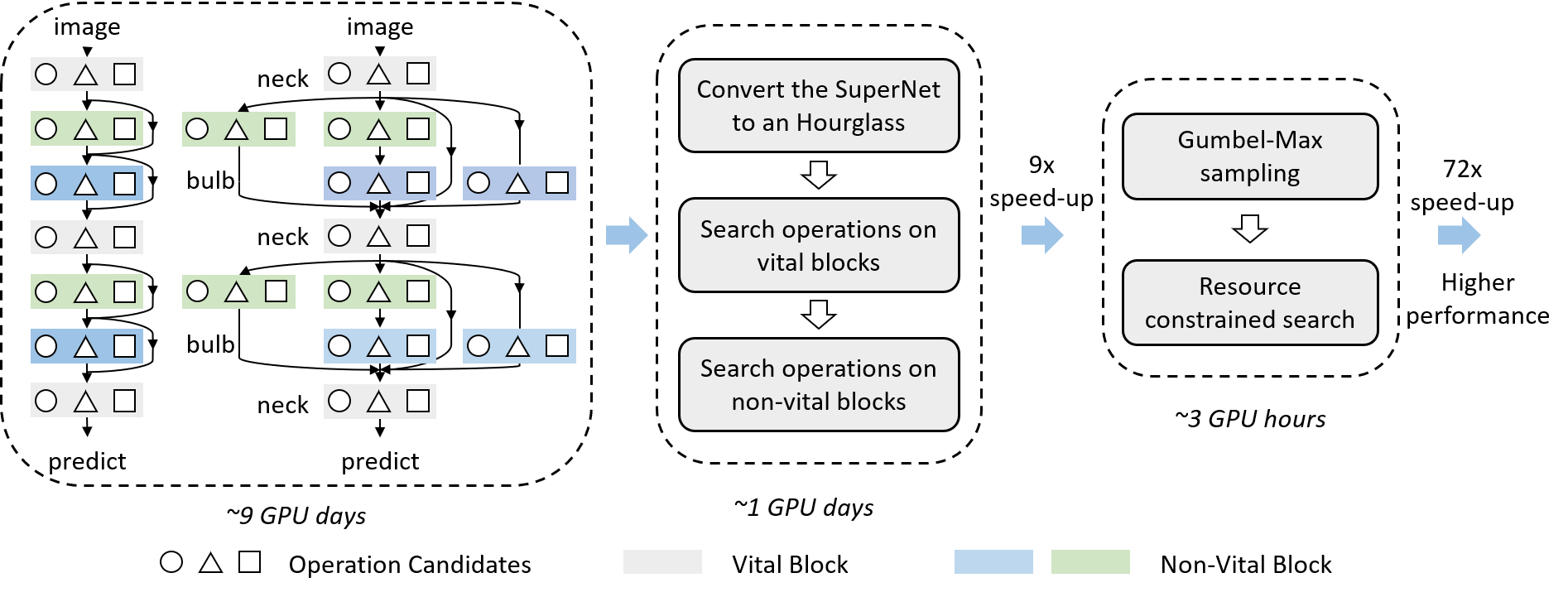}}
	\caption{Blocks in the residual network are either ``vital'' or ``non-vital'', and they form the neck or bulb parts in the hourglass network. Two-stage search scheme speed up architecture search by 9$\times$ and the resource constrained search further accelerates architecture search by 72$\times$.}
	\label{fig_HourNAS}
	\vspace{-10pt}
\end{figure*}

These methods have made tremendous efforts to greatly accelerate the search process of NAS. Nevertheless, given the huge computation cost on the large-scale dataset~\cite{fbnet,sposnas,resourceconstrainednas,amc,onceforall}, \emph{e.g.}, 9 GPU days for NAS on the ImageNet benchmark, most methods execute NAS in a compromised way.  The architecture is first searched on a smaller dataset (\emph{e.g.}, CIFAR-10 ~\cite{cifar}), and then the network weight of the derived architecture is trained on the large dataset. An obvious disadvantage of this concession is that the performance of the selected architecture on the CIFAR-10 may not be well extended to the ImageNet benchmark~\cite{undersandingrobustfyingnas}. We tend to use the minimal number of parameters without discrimination to construct an architecture that would achieve the maximal accuracy. But just as the popular 80-20 rule\footnote{{\color{black} The 80/20 rule (a.k.a Pareto principle, the law of vital few) is an aphorism that states, for many events, roughly 80\% of the effects come from 20\% of the causes.}} goes, only a few parts could be critical to the architecture's success, and we need to give them the most focus while balancing the parameter volume for other parts.

In this paper, we propose HourNAS for an accurate and efficient architecture search on the large-scale ImageNet dataset. Blocks in an architecture are not created equally. Given all the possible paths for the information flow from the input to the output of the network, blocks shared by these paths are \emph{vital}, just as the neck of an hourglass to regulate the flow. We identify these vital blocks and make them the priority in the architecture search. The other non-vital blocks may not be critical to the accuracy of the architecture, but they often occupy the major volume of the architecture (like the bulb of an hourglass) and will carve up the resource budget left by the vital blocks. Instead of directly working in a large search space flooding with architectures that obviously do not comply with the constraints during deployment, we develop a space proposal method to screen out and optimize the most promising architecture candidates under the constraints. By treating the architecture search through an hourglass lens, the limited computation resource can be well allocated across vital and non-vital blocks. We design toy experiments on residual networks to illustrate the varied influence of vital and non-vital blocks on the network accuracy. The resulting HourNAS costs only about 3 hours~(0.1 GPU days) on the entire ImageNet dataset to achieve a 77.0\% Top-1 accuracy, which outperforms the state-of-the-art methods.

\section{Related Works}\label{sec:relatedworks}

This section reviews the methods for neural architecture search algorithms. Then, we discuss layer equality, \emph{i.e.}, the importance of different blocks in deep neural networks.

\iffalse
{\noindent \bf Efficient Network Design} 
In early age of deep learning, most widely used neural architectures are manually designed by human experts, \emph{e.g.}, VGG~\cite{vgg}, ResNet~\cite{resnet}, ~\cite{densenet}, and \emph{etc}. With the increasing demand for real-time applications on mobile or embedded devices, designing efficient neural networks has attracted more research interest. MobileNet serials~\cite{mobilenet, mobilenetv2} utilize the depthwise separable convolution and inverted residual block to construct efficient networks. ShuffleNet serials~\cite{shufflenet,shufflenetv2} employ pointwise group convolution and channel shuffle module to design computation-efficient models for mobile devices. Despite their strong performance, manually designed networks require a huge effort of human experts.
%MobileNetV1~\cite{mobilenet} utilizes depthwise separable convolution to reduce the computation cost drastically. MobieNetV2~\cite{mobilenetv2} further improves efficiency by introducing a novel inverted residual module with a linear bottleneck. ShiftNet~\cite{shift} was proposed to reduce the computation cost of conventional convolutional networks by a novel FLOP-free and parameter-free shift operation. 
\fi

{\noindent \bf Neural Architecture Search.} To automate the design of neural models, neural architecture search (NAS) was introduced to discover efficient architectures with competitive performance. Reinforcement learning (RL) and evolution algorithm~(EA) were widely adopted in NAS~\cite{nasnet,zoph_rl_iclr,nsganet,amoebanet,mnasnet,sposnas,autogan}. However, these methods were highly computationally demanding. Many works have been devoted to improving the efficiency of NAS from different perspectives, \emph{e.g.}, by adopting the strategy of weight sharing~\cite{enas} or progressive search~\cite{pnas}. Differentiable based NAS attracted great interest as it drastically reduces the searching cost to several days or even hours~\cite{fbnet,darts,setnas,snas,autoreid,autodeeplab}. For example, DARTS~\cite{darts} adopted the continuous architecture representation to allow efficient architecture search via gradient descent. {\color{black} Meanwhile, as discusses in TuNAS~\cite{tunas}, gradient-based NAS methods consistently performed better than random search, showing the power of searching for excellent architectures. However, the most efficient gradient-based NAS methods still took dozens of days for directly searching on target large-scale dataset (\eg, ImageNet).} Thus, an efficient method for directly searching deep neural architectures on large-scale datasets and search spaces is urgently required. 

%\emph{e.g.}, MobileNetV3~\cite{mobilenetv3} was obtained by hardware aware neural architecture search and presents stronger performance than its counterparts (MobileNetV1~\cite{mobilenet} and MobileNetV2~\cite{mobilenetv2}).   
%A similar idea of differential NAS was also adopted in~\cite{snas,fbnet} and equipped with Gumbel softmax. Despite the strong performance and high efficiency of differential NAS methods, there are still some problems that are not thoroughly explored, \emph{i.e.}, which paths in the network are more important for the final performance and thus need to be emphasized in the search process. 

{\noindent \bf Layer Equality.} Most of existing NAS methods treated all layers with the same importance during the search. However, convolution neural networks are always over-parameterized and the impact on the final accuracy of each layer is totally different. Zhang \emph{et al.}~\cite{arealllayerscreatedequal} re-initialized and re-randomized the pre-trained networks, and found that some layers are robust to the disturbance. For some intermediate layers, the re-initialization and re-randomization steps did not have negative consequences on the accuracy. Veit \emph{et al.}~\cite{resnetensembles} decomposed residual networks and found that skipping some layers does not decrease the accuracy significantly. Ding \emph{et al.}~\cite{globalsparsemomentumsgd} pruned different layers separately and found some layers are more important to the final accuracy. It is obvious that the layers are not created equal, and some layers are more important than others. 

In this paper, we analyze the causes of the inequality phenomenon in the residual network and exploit this property in neural architecture search to improve its efficiency.

\section{Hourglass Neural Architecture Search}
In this section, we revisit the neural architecture search from an hourglass way. The vital few blocks should be searched with a higher priority to guarantee the potential accuracy of the architecture. The non-vital blocks that occupy the major volume are then searched in an extremely fast way by focusing on the discovered space proposals.

\subsection{Vital Blocks: the Neck of Hourglass}\label{sec_notequal}
In this paper, we focus on the serial-structure NAS SuperNet~\cite{mnasnet, efficientnet, fbnet, proxylessnas}, as it is hardware-friendly and capable of achieving superior performance. Before we illustrate vital blocks in a general NAS, we first take ResNet~\cite{resnet} as an example for the analysis. ResNet is one of the most popular manually designed architectures. It is established by stacking a series of residual blocks, and the residual block is defined as,
\begin{equation}
\mathbf{y} = \mathcal{F}(\mathbf{x}, \mathbf{w}) + \mathbf{x},
\label{eq_resnet}
\end{equation}
where $\mathbf{x}$ is the input feature map, $\mathcal{F}$ denotes the transformation (\emph{e.g.}, convolution and batch normalization for vision tasks) and $\mathbf{w}$ stands for the trainable parameters.\footnote{As for the downsample blocks (reduce the feature map size) and the channel expansion blocks (increase the channel number), we follow~\cite{resnetensembles} and use $\mathbf{y} = \mathcal{F}(\mathbf{x}, \mathbf{w})$ to express.} From the information flow perspective, there are two paths to transmit the information from the node $\mathbf{x}$ to the node $\mathbf{y}$, \emph{i.e.}, the shortcut connection and the transformation $\mathcal{F}$. If there are $m$ residual blocks in a network, there will be $2^m$ different paths for the information propagation in total. A general neural network $\mathcal{N}$ based on the residual blocks~\cite{resnet, mnasnet, fbnet} can therefore be approximated as the ensemble of a number of paths~\cite{resnetensembles} $\{\mathcal{P}_1, \dots, \mathcal{P}_n\}$, \emph{i.e.}, $\mathcal{N}(X) \approx \sum_{i=1}^{n} \mathcal{P}_i (X)$,
where each path $\mathcal{P}_i$ is set up by a series of blocks, $X$ is the input data, and $n$ is the number of all the paths. 

It is worth noticing that there are a few blocks existing in all the possible paths, \emph{e.g.}, the gray blocks in Fig.~\ref{fig_HourNAS}. These self-contained blocks do not participate in forming any residual blocks, but they are \emph{vital}, because of their appearance in every path from the input to the output of the network. On the other hand, the green and blue blocks in Fig.~\ref{fig_HourNAS} are a part of the residual blocks $\mathbf{y} = \mathcal{F}(\mathbf{x}, \mathbf{w}) + \mathbf{x}$, where the information can be transmitted through the plain transformation $\mathcal{F}(\mathbf{x}, \mathbf{w})$ or the shortcut connection $\mathbf{x}$ to the next block, {\color{black} so they are not that vital}. 

{\color{black} 
\noindent \textbf{Identify and Examine Vital Blocks.}} Given the paths $\{\mathcal{P}_1, \dots, \mathcal{P}_n\}$ in a general residual network $\mathcal{N}$, the vital blocks shared by all the paths can be identified through $\hat{\mathcal{P}} = \mathcal{P}_1 \cap \dots \cap \mathcal{P}_n$, where $\mathcal{P}_i \cap \mathcal{P}_j$ denotes the intersection set of those blocks in paths $\mathcal{P}_i$ and $\mathcal{P}_j$. In the popular residual networks, such as ResNet~\cite{resnet} and FBNet~\cite{fbnet}, the vital blocks are exactly the first convolution layer, the last fully connected layer, the downsampling blocks, and the channel expansion blocks. These vital blocks are critical to the accuracy of the whole architecture, as they exist in all paths and act as the neck of the hourglass to control the information flow. In contrast, the other blocks would always find substitutes for themselves to keep the information flow, and they thus play a secondary role in the whole architecture. 

We further take mobile architectures as an example to illustrate the different influence of vital and non-vital blocks on the network accuracy. A random mask function $\mathcal{M}(\mathbf{y}, p)$ is introduced to destroy the output of blocks in the pretrained MnasNet~\cite{mnasnet} and MobileNetV2~\cite{mobilenetv2}, where $\mathbf{y}$ is the output feature map, and $0\le p\le 1$ is the  probability. In particular, every channel is reset to 0 with a probability $p$. 
\begin{figure}
	%\
	\centering
	\subfigure[MnasNet]{\includegraphics[width=0.48\linewidth]{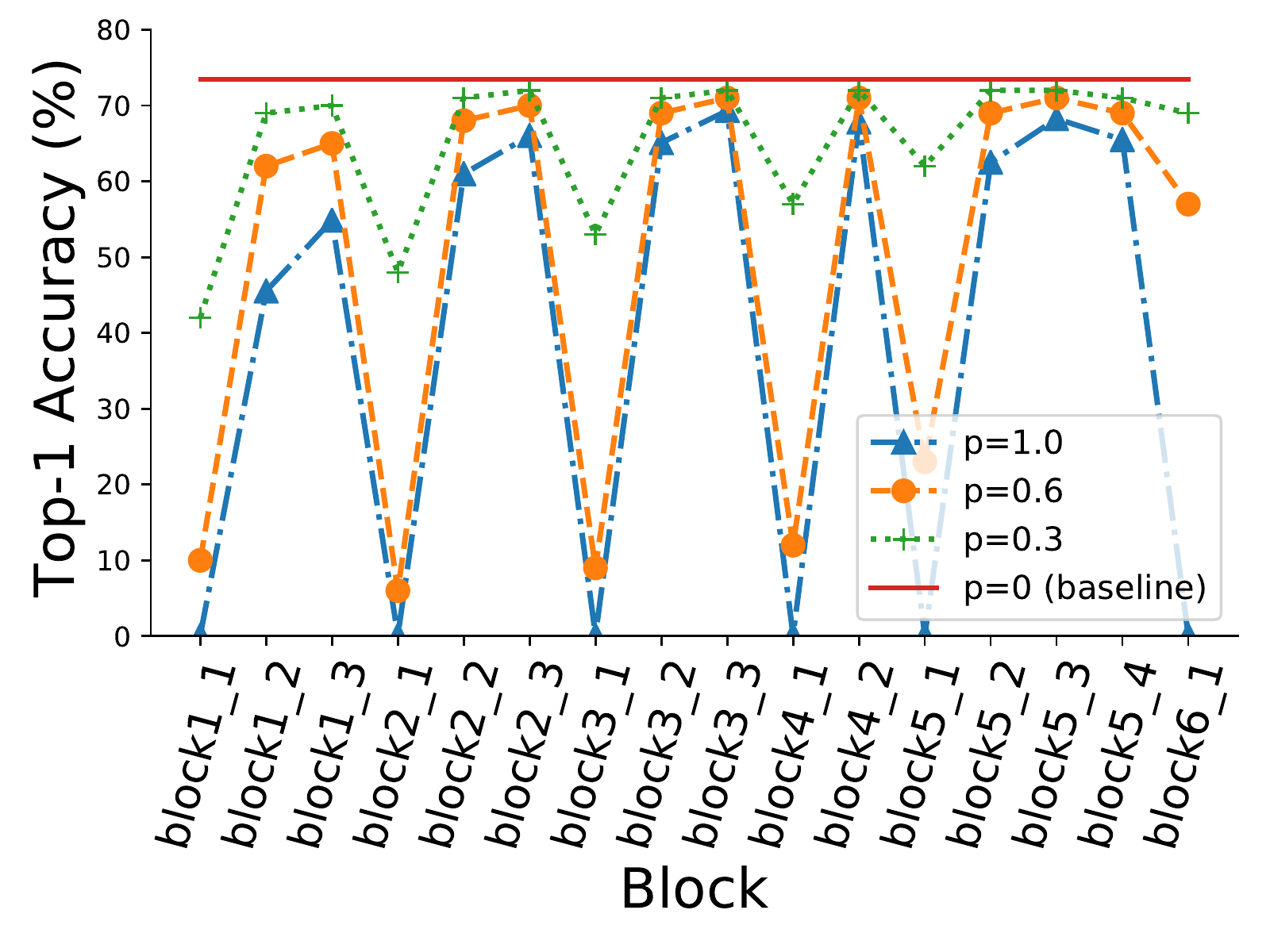}}
	\subfigure[MobileNetV2]{\includegraphics[width=0.48\linewidth]{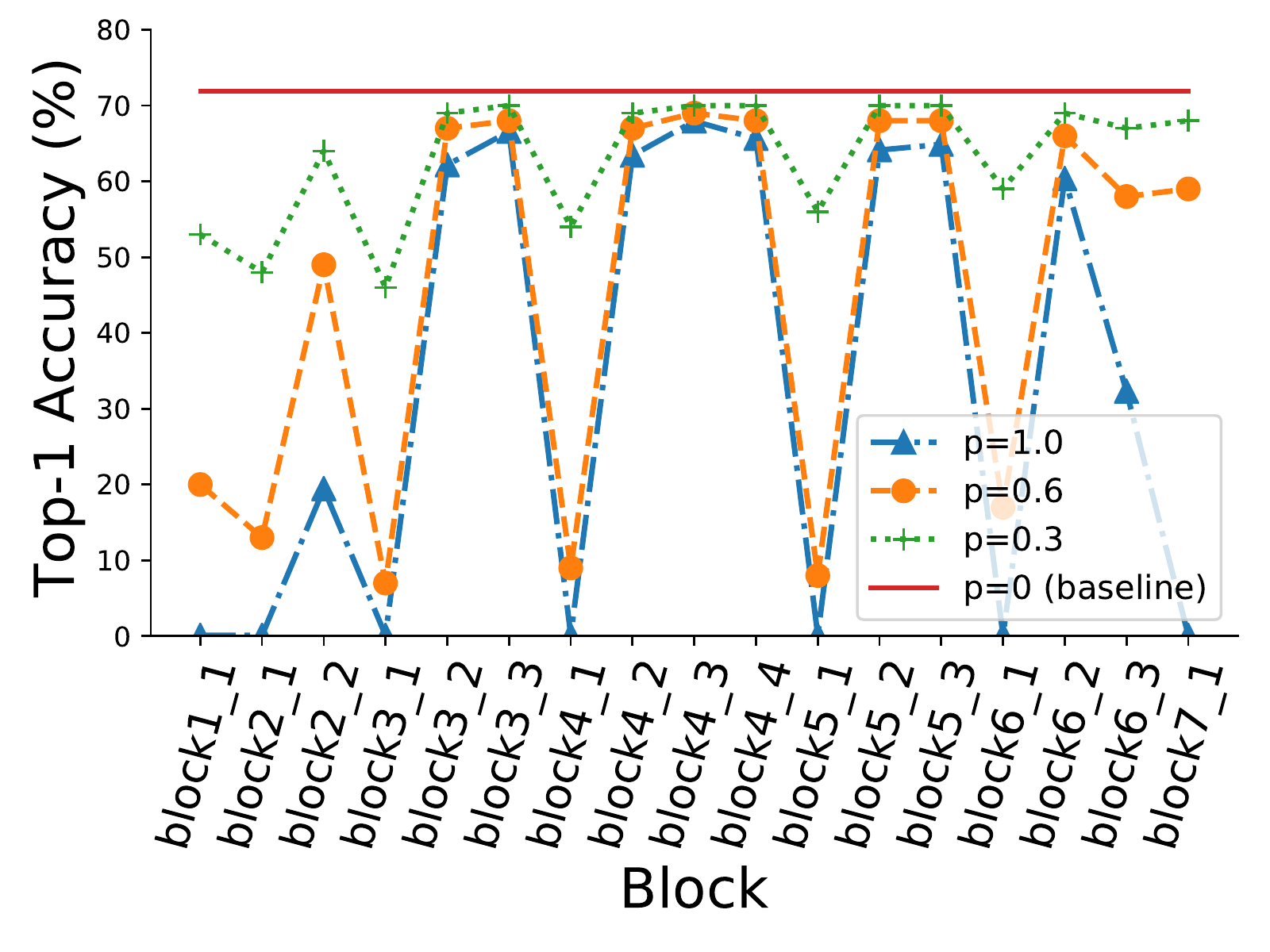}}
	%\
	\caption{The diagram of block importance by using the MnasNet and MobileNetV2 pretrained models.}
	\label{fig_block_importance}
	\vspace{-15pt}
\end{figure}
Fig.~\ref{fig_block_importance} shows the accuracy change of MnasNet~\cite{mnasnet} and MobileNetV2~\cite{mobilenetv2} resulting from the feature distortion on different blocks. {\color{black} The $blocki\_j$ denotes the $j$-th block in the $i$-th stage. As discussed above, the first block in every stage is the vital block}. We set $p = \{0.3, 0.6, 1.0\}$ to gradually increase the degree of feature distortion. Each time we only manipulate one block while keeping others unchanged in the network. For the non-vital blocks (\emph{e.g.}, {\color{black}block3\_2 and block3\_3 in both MnasNet and MobileNetV2}), even if all the channels are reset to zero (\emph{i.e.}, p=1.0), the network does not undergo a significant accuracy degradation. However, a small portion (p=0.3) of channels that are masked out for those vital blocks ({\color{black}\emph{e.g.}, block1\_1 and block2\_1}) will lead to an obvious accuracy drop.

\iffalse
{\color{black}
	\noindent \textbf {Theoretical Analysis of Vital blocks.} In this part, we try to analyze why some layers play a more vital role by analyzing the rank. The rank in the feature represents the richness of information, and also represents the distinctiveness~\cite{yang2017breaking}. We start by considering two transformations,
	\begin{equation}
	\begin{aligned}
	y_1 &= \mathcal{F}(w \otimes x),\\
	y_2 &= \mathcal{F}(w \otimes x) + x,
	\end{aligned}
	\end{equation}
	where $\mathcal{F}$ is the non-linear transformation, $\otimes$ is the matrix multiplication and $x$, $y$ are the input and output, respectively. The ranks of the outputs are:
	\begin{equation}
	\begin{aligned}
	rank(y_1) &\le \min\{rank(w), rank(x)\}, \\
	rank(y_2) &\le \min\{rank(w), rank(x)\} + rank(x).
	\end{aligned}
	\end{equation}
	Adding the identity mapping makes $y_2$ capable of containing much more information, and the high rank feature maps are critical to the final performance~\cite{hrank}. The non-vital blocks and the vital blocks act like $y_2$ and $y_1$, respectively. For example, removing the learned transformation $\mathcal{F}$ of these transformations does not affect much to the whole network as Fig.~\ref{fig_block_importance} shows because the information could be preserved from the shortcut. For the transformation $y_1 = \mathcal{F}(w \otimes x)$, this is how downsampling blocks transform features, the quality of the transformation $\mathcal{F}$ itself determines the information preservation.
}
\fi

{\color{black} \noindent \textbf{Revisit Neural Architecture Search.}} The goal of NAS is to search for the architecture of a higher accuracy under the constraints of available computational resources. In other words, NAS can be regarded as a problem of resource allocation. Vital blocks are potentially the most important and need to be put as the priority. As a result, more resources are better to be first allocated to the vital blocks, and the remaining resources are used for the non-vital blocks design. This therefore naturally motivates us to develop a two-stage search algorithm. During the first stage, we construct the minimal SuperNet $\mathcal{S}_{vital}$ {\color{black} by stacking all the vital layers and} search the vital blocks. {\color{black} The weights and architecture parameters are optimized alternatively in a differentiable way~\cite{fbnet}.} In the second stage, we fix the derived architecture of those vital blocks, and allocate the computational resources to search for the non-vital blocks.

\subsection{Non-Vital Blocks: the Bulb of Hourglass}

Non-vital blocks are often composed of a large number of parameters. They look like the bulb of the hourglass to determine the whole volume size. If the computational resources are unlimited to deploy the neural network, then we can make the network wider and deeper for achieving a higher performance~\cite{resnet, efficientnet}. However, the searched architectures are to be deployed on mobile devices which have demanding constraints of the resources, \emph{e.g.}, model size, and FLOPs. Without investigating these constraints, it would be ineffective to directly sample the architecture from a large search space. For example, if the sampled architectures cannot fully use the available computation resource, the resulting models might perform poorer than expected, which has been analyzed by a number of multi-objective NAS works~\cite{nsganet, cars, lemonade, mnasnet} (the Pareto front). Otherwise, if the sampled architectures overwhelm the use of computation resources,  they would be difficult to be deployed. To tackle the dilemma, we introduce an efficient sampling operation to {\color{black} avoid wasting too much time on search unimportant operations}, and a space proposal strategy to put more attention on {\color{black} architectures that meet the target computational resources}.

\subsection{Space Proposal for Efficient Resource Constrained Search}
\label{sec_space_proposal}
A general differentiable neural architecture search algorithm can be formulated as a bilevel optimization problem~\cite{darts}:
\begin{equation}
\begin{aligned}
\theta^* &= \mathop{\mathrm{\arg\min}}_{\theta} \mathcal{H}_{val}(w^*(\theta), \theta),\\
\textnormal{s.t.} &~~ w^*(\theta) = \mathop{\mathrm{\arg\min}}_{w} \mathcal{H}_{train}(w, \theta),
\end{aligned}
\end{equation}
where $\mathcal{H}$ is the cross-entropy loss function, and $\theta$ denotes the architecture parameters. If the accuracy is the only objective to be considered for searching, a complex architecture would be preferred for achieving a highest accuracy (see Sec.~\ref{sec_ablation}). However, if the obtained architectures are to be deployed on mobile devices, we may always have the computational resource constraints from the environment . Thus, neural architecture search that considers the multiple objectives can be formulated as,
\begin{equation}
\begin{aligned}
\theta^* &= \mathop{\mathrm{\arg\min}}_{\theta} \mathcal{H}_{val}(w^*(\theta), \theta) + \alpha \times \mathcal{T}(\theta),\\
\textnormal{s.t.} &~~ w^*(\theta) = \mathop{\mathrm{\arg\min}}_{w} \mathcal{H}_{train}(w, \theta),
\end{aligned}
\end{equation}
where $\mathcal{T}(\theta)$ is the regularization term that encourages the produced architecture to satisfy the target computational resource constraints. Assuming the constraints~(targets) on computational resources (\eg, model size, FLOPs) are $T_{i \in \{1, \dots, n\}}$, where $n$ is the number of objectives, an efficient and controllable way is to initialize architectures that satisfy $T$. Thus, we introduce the concept of space proposal. The space proposal is a subspace of the large search space, and all the sampled architectures from the space proposal satisfy the target resources. As a result, the search phase would not waste resources on optimizing useless architectures. In addition, the space proposal ensures ``what you set is what you get''. Similar to gradient-based NAS, the space proposal is represented by the architecture parameters.

We take the FLOPs as an example to describe how to optimize a space proposal. Suppose $\theta$ represents the trainable architecture parameters of the NAS SuperNet and the size is $L\times O$, where $L$ is the maximum depth and $O$ is the number of candidate operations {\color{black} in each layer}. A number of methods $\mathcal{G}$ are capable of sampling architectures $A_{\theta}$ from architecture parameters $\theta$,
\begin{equation}
A_{\theta} = \mathcal{G}(\theta),~~\sum_{o} A_{l,o} = 1,
\label{eq_sample}
\end{equation}
where $\mathcal{G}$ is usually specified as softmax~\cite{darts}, Gumbel-softmax~\cite{fbnet}, Gumbel-Max~\cite{gdas,data}, \emph{etc}. The $A_{l,o}$ is the $o$-th operation in the $l$-th layer.

The FLOPs table $F$ of the SuperNet $\mathcal{S}$ is of size $L \times O$, where $F_{l, o}$ denotes the FLOPs of the $o$-th operation in the $l$-th SuperBlock. The FLOPs for sampled architecture $A_\theta$ is calculated as $\mathcal{R}_{F}(A_\theta) = \textnormal{sum} (A_\theta \odot F)$, where $\odot$ is the element-wise product. Assuming the target FLOPs is $T_{F}$, the optimization is formulated as,
\begin{equation}
\theta^F = \mathop{\mathrm{\arg\min}}_{\theta} | \mathcal{R}_{F}(\mathcal{G}(\theta)) - T_{F} | / M_{F},
\label{eq_clock_loss_flops}
\end{equation}
where $M_{F}$ is a constant scalar denotes the maximum FLOPs of the sampled architectures, and this term is used for normalizing the objective to $[0, 1]$. We extend Eqn~\ref{eq_clock_loss_flops} to $n$ different objectives. The targets for $n$ objectives are $T_{i \in \{1, \dots, n\}}$, and the optimization is defined as,
\begin{equation}
\theta^T = \mathop{\mathrm{\arg\min}}_{\theta} \frac{1}{n}\sum_{i=1}^n | \mathcal{R}_i (\mathcal{G}(\theta)) - T_i | / M_i,
\label{eq_clock_loss}
\end{equation}
which $\mathcal{R}_i (\mathcal{G}(\theta))$ is the resource demand of the architecture sampled by $\mathcal{G}(\theta)$ on the $i$-th objective. 

This optimization problem is easily to be solved in a few seconds. The solution $\theta^T$ can be regarded as a space proposal under the constraints $T$, and the structure $A_{\theta^T}$ sampled from $\theta^{T}$ by Eqn~\ref{eq_sample} would be more easily to satisfiy target resources $T$. Instead of relying on a single optimal solution $\theta^T$, we turn to an ensemble way to start from different random initializations and derive a series of space proposals $\Theta^{T} = \{ \theta^{T}_1, \cdots, \theta^{T}_m \}$, where $m$ is the number of space proposals. The orthogonal constraint is also introduced to further increase the diversity of different space proposals, which is formulated as,
\
\begin{equation}
\begin{aligned}
\theta_1, \dots, \theta_m = \mathop{\arg\min}_{\theta_1, \dots, \theta_m} (\frac{1}{nm}&\sum_{j=1}^m \sum_{i=1}^n |\mathcal{R}_i(\mathcal{G}(\theta_j)) - T_i| / M_i + \\ &\beta \times \sum (|O - I|)),
\end{aligned}
\label{eq_regularization_loss}
\end{equation}
where $O = \Theta\Theta^T$ is an $m\times m$ matrix and element $O_{i, j}$ denotes the inner product of $\theta_i$ and $\theta_j$. The term $\sum(|O-I|)$ regularizes $m$ space proposals to be orthogonal, which indicates that the architectures sampled from different space proposals are different. A uniformly initialized auxiliary parameter $\Pi$ of size $m$ is then introduced to sample space proposals,
\
\begin{equation}
\pi = \mathcal{P}(\Pi),~~\sum_i \pi_i = 1,
\label{eq_pi}
\end{equation}
where $\mathcal{P}$ could be softmax, Gumbel-softmax or Gumbel-Max, $\pi$ is the sampled vector from $\Pi$ that used for combine the architectures sampled from $m$ space proposals, and the ensembled architecture $A_{\Theta}$ that used for updating the SuperNet is defined as,
\
\begin{equation}
A_{\Theta} = \sum_{i} \pi_i \cdot A_{\theta_i} = \sum_{i} \pi_{i} \cdot \mathcal{G}(\theta_i),
\end{equation}
where $A_{\Theta}$ is utilized for updating network parameter $w$ on train set $\mathcal{D}_{train}$ and architecture parameter $\Pi, \Theta$ on validation set $\mathcal{D}_{val}$, respectively. Every space proposal $\theta_i$ optimizes towards the good architectures in the space proposal and $\Pi$ optimizes towards better proposals. The NAS framework by using the space proposal strategy is summarized as,
\
\begin{equation}
\begin{aligned}
\Pi^*, \Theta^* &= \mathop{\mathrm{\arg\min}}_{\Pi, \Theta} \mathcal{H}_{val}(w^*(\Pi, \Theta), \Pi, \Theta) + \alpha \times \mathcal{T}(\Pi, \Theta), \\
\textnormal{s.t.} &~~ w^*(\Pi, \Theta) = \mathop{\mathrm{\arg\min}}_{w} \mathcal{H}_{train}(w, \Pi, \Theta),
\end{aligned}
\end{equation}
where $\mathcal{T}$ (Eqn~\ref{eq_clock_loss}) is the regularization on space proposal parameters (Eqn~\ref{eq_regularization_loss}), and $\alpha$ is the slope of the multi-objective loss term. 

\subsection{Overall Search Algorithm}

{\color{black} Based on the proposed method, we summarize the overall search algorithm in Alg.~\ref{alg_fnas}.}
\begin{algorithm}[h]
	\caption{The searching algorithm of HourNAS.}
	\begin{algorithmic}[1]
		\REQUIRE The NAS supernet $\mathcal{S}$, the computational targets $T_{i \in \{1, \dots, n\}}$, the train set $D_{train}$ and validation set $D_{val}$, the searching epochs for vital blocks $E_{vital}$ and non-vital blocks $E_{n-vital}$, the number of space proposals $m$, iterations $I_{sp}$ for training space proposals.
		\STATE \textbf{// Search Vital Blocks} 
		\STATE Constructing the minimal SuperNet $\mathcal{S}_{vital}$ {\color{black} by stacking all the vital layers} and the architecture parameter $\theta_{vital}$.
		\FOR{e = 1 to $E_{vital}$}
		\FOR{data and target pair $(X_{tr}, Y_{tr})$ in $D_{train}$}
		\STATE Sample network $A$ from $\theta_{vital}$, calculate loss and update network parameters.
		\ENDFOR
		\FOR{data and target pair $(X_{val}, Y_{val})$ in $D_{val}$}
		\STATE Sample network $A$ from $\theta_{vital}$, calculate loss and update $\theta_{vital}$. 
		\ENDFOR
		\ENDFOR
		\STATE The operations with the highest importance are selected to form the vital layers.
		\STATE \textbf{// Optimize $m$ space proposals}
		\STATE According to the computational targets $T$, HourNAS optimizes $m$ proposals $\Theta^{T} = \{ \theta^{T}_1, \cdots, \theta^{T}_m \}$ for $I_{sp}$ iterations (Eqn~\ref{eq_regularization_loss}), and construct the proposal sampler $\pi$~(Eqn~\ref{eq_pi}).
		\STATE \textbf{// Search Non-Vital Blocks}
		\FOR{e = 1 to $E_{n-vital}$}
		\FOR{data and target pair $(X_{tr}, Y_{tr})$ in $D_{train}$}
		\STATE Sample network $A$ from $\pi$ and $\Theta$, calculate loss and update network parameters.
		\ENDFOR
		\FOR{data and target pair $(X_{val}, Y_{val})$ in $D_{val}$}
		\STATE Sample network $A$ from $\pi$ and $\Theta$, calculate loss and update $\pi$ and $\Theta$.
		\ENDFOR
		\ENDFOR
		\STATE Fix operations by selecting the space proposal and operations with the highest probability.
		\ENSURE The architecture $A$ which satisfies the computational targets $T_{i \in \{1, \dots, n\}}$.
	\end{algorithmic}
	\label{alg_fnas}
\end{algorithm}

\section{Experiments}\label{sec:exp}
In this section, we first describe the experimental settings and then extensively evaluate the proposed HourNAS on several popular NAS search spaces~\cite{fbnet, mnasnet, efficientnet} on ImageNet.

\subsection{Experimental Settings}
We use the HourNAS to search on the complete ImageNet train set. The subset $\mathcal{D}_{tr}$ takes $80\%$ of the train set to train network parameters and the rest $\mathcal{D}_{val}$ is used to update architecture parameters. We search on three popular search spaces, \ie, FBNet~\cite{fbnet}, MnasNet~\cite{mnasnet}, and EfficientNet~\cite{efficientnet}. For any of our searched architecture, the training strategy is the same as the corresponding baseline. {\color{black} We use the NVIDIA V100 GPU to measure the search time and compare with previous works fairly. The V100 GPU is also adopted by a number of literatures, for example, PDARTS~\cite{pdarts}, FBNet~\cite{fbnet}.}

The HourNAS first searches the vital blocks for one epoch~(about 1 hour). Then, HourNAS optimizes multiple space proposals according to the computational targets and searches the non-vital blocks for one epoch~(about 2 hours). The whole searching process requires only one V100 GPU within 3 hours. Extending the search time will not further improve the accuracy, because the distribution of architecture parameters is stable. For competing methods like MnasNet~\cite{mnasnet}, 3 GPU hours could only train one sampled architecture and the RL controller has no difference with random search. We utilize the Gumbel-Max~\cite{astarsampling, data, gdas} to sample operations according to the learned importance, which avoids wasting searching time on undesired operations. Gumbel-Max samples an operation according to the learned probability distribution (\emph{i.e.}, importance). The sampling frequencies of those poor operations tend to be relatively low, so that we can effectively reduce the time spent on them. The Gumbel-Max sampling accelerates every iteration by around $O$ times, where $O$ is the number of candidate operations in every layer. During searching, we follow FBNet~\cite{fbnet} and add the temperature $\tau$ (Eqn~\ref{eq_sample}) for sharpening the distribution progressively. The temperature $\tau$ starts from 5.0 and multiply 0.9999 at every iteration. Slope parameter $\alpha$, $\beta$ are emperically set to 5.0 and 1e-2, respectively. Learning rates for optimizing weights and architecture parameters are 0.1 and 0.01, respectively. Adam~\cite{adam} optimizer is used to update architecture parameters.

\begin{table*}
	\caption{Overall comparison on the ILSVRC2012 dataset.}
	\small
	\centering
	%\begin{adjustwidth}{-6pt}{}
	\begin{tabular}{c|c|c|c|c|c|c|c}
		\toprule
		\multirow{2}{*}{{Model}} & \multirow{2}{*}{{Type}} & Search & Search Cost & Params & FLOPS & Top-1 & Top-5 \\
		& & Dataset & (GPU days) & (M) & (M) & (\%) & (\%) \\
		\midrule
		ResNet50~\cite{resnet} & manual & - & - & 25.6 & 4100 & 75.3 & 92.2 \\
		MobileNetV1~\cite{mobilenet} & manual & - & - & 4.2 & 575 & 70.6 & 89.5 \\
		MobileNetV2~\cite{mobilenetv2} & manual & - & - & 3.4 & 300 & 72.0 & 91.0 \\
		MobileNetV2~(1.4$\times$) & manual & - & - & 6.9 & 585 & 74.7 & 92.5 \\
		ShuffleNetV2~\cite{shufflenetv2} & manual & - & - & - & 299 & 72.6 & - \\
		ShuffleNetV2~(1.5$\times$) & manual & - & - & 3.5 & 299 & 72.6 & - \\
		\midrule
		FPNASNet~\cite{fpnas} & auto & CIFAR-10 & 0.8         & 3.4 & 300 & 73.3 & - \\ % Original
		SNAS~(mild)~\cite{snas} & auto & CIFAR-10 & 1.5       & 4.3 & 522 & 72.7 & 90.8\\ % PDARTS
		AmoebaNet-A~\cite{amoebanet} & auto & CIFAR-10 & 3150 & 5.1 & 555 & 74.5 & 92.0 \\ % PDARTS
		PDARTS~\cite{pdarts} & auto & CIFAR-10 & 0.3          & 4.9 & 557 & 75.6 & 92.6 \\ % PDARTS
		NASNet-A~\cite{nasnet} & auto & CIFAR-10 & 1800       & 5.3 & 564 & 74.0 & 91.3 \\ % PDARTS
		GDAS~\cite{gdas} & auto & CIFAR-10 & 0.2              & 5.3 & 581 & 74.0 & 91.5 \\
		PNAS~\cite{pnas} & auto & CIFAR-10 & 225              & 5.1 & 588 & 74.2 & 91.9 \\ % DARTS
		CARS-I~\cite{cars} & auto & CIFAR-10 & 0.4            & 5.1 & 591 & 75.2 & 92.5 \\
		DARTS~\cite{darts} & auto & CIFAR-10 & 4              & 4.9 & 595 & 73.1 & 91.0 \\ % PDARTS
		MdeNAS~\cite{mdenas} & auto & CIFAR-10 & 0.2          & 6.1 & - & 74.5 & 92.1 \\ % Oringinal
		%SPNAS~\cite{spnas} & auto & ImageNet & 32 & - & - & 75.0 & - \\
		RCNet~\cite{resourceconstrainednas} & auto & ImageNet & 8     & 3.4 & 294 & 72.2 & 91.0 \\ % Original
		%Once-For-All~\cite{onceforall,bignas} & auto & ImageNet & 1.7 & 4.4 & 327 & 75.3 & - \\
		SPOSNAS~\cite{sposnas} & auto & ImageNet & 13                 & 5.3 & 465 & 74.8 & - \\ % SPOSNAS
		ProxylessNAS~\cite{proxylessnas} & auto & ImageNet & 8.3      & 7.1 & 465 & 75.1 & 92.5 \\ % APDARTS
		%MobileNetV3~\cite{mobilenetv3} & auto & xxx & xxx & 5.4 & 217 & 75.2 & - \\
		\midrule
		FBNet-B~\cite{fbnet} & auto & ImageNet & 9 & 4.8 & 295 & 74.1 & - \\ % SPOSNAS
		FBNet-C~\cite{fbnet} & auto & ImageNet & 9 & 5.5 & 375 & 74.9 & - \\ % SPOSNAS
		\textbf{HourNAS-FBNetSS-A} & auto & ImageNet & \textbf{0.1} & 4.8 & 298 & \textbf{74.1} & \textbf{91.8} \\
		\textbf{HourNAS-FBNetSS-B} & auto & ImageNet & \textbf{0.1} & 5.5 & 406 & \textbf{75.0} & \textbf{92.2} \\
		\midrule
		\textbf{HourNAS-EFBNetSS-C} & auto & ImageNet & \textbf{0.1} & 4.8 & 296 & \textbf{74.1} & \textbf{91.6} \\
		\textbf{HourNAS-EFBNetSS-D} & auto & ImageNet & \textbf{0.1} & 5.5 & 394 & \textbf{75.3} & \textbf{92.3} \\
		\midrule
		MnasNet-A1~\cite{mnasnet} & auto & ImageNet & 3800 & 3.9 & 312 & 75.2 & 92.5 \\ %FBNet
		\textbf{HourNAS-MnasNetSS-E} & auto & ImageNet & \textbf{0.1} & 3.8 & 313 & \textbf{75.7} & \textbf{92.8} \\
		\midrule
		EfficientNet-B0~\cite{efficientnet} & auto & ImageNet & - & 5.3 & 390 & 76.8 & - \\
		\textbf{HourNAS-EfficientNetSS-F} & auto & ImageNet & \textbf{0.1} & 5.3 & 383 & \textbf{77.0} & \textbf{93.5} \\
		\bottomrule
	\end{tabular}
	%\end{adjustwidth}
	\label{tab_imagenet}
	%\vspace{-15pt}
\end{table*}

\subsection{Comparison with State-of-the-arts}
\textbf{FBNet Search Space (FBNetSS).} We first evaluate our HourNAS on the popular FBNet search space (FBNetSS), {\color{black} which contains $9^{22} \approx 1\times 10^{21}$ architectures}. HourNAS first searches the vital blocks and the results show that operations with the expansion ratio of six have significantly higher probabilities than other operations. We choose the operations with the highest probabilities to form the vital blocks. This result is in line with our intuition that the complex operations have the greatest feature extraction ability on the large dataset, \emph{i.e.}, ImageNet.

After fixing the operations of the vital blocks, we are interested in finding the appropriate proposal number $m$. We set the computational resources the same as FBNet-B and search architectures using $m$ different space proposals. We first visualize the distribution of sampled architectures from the space proposal by gradually decreasing the temperature $\tau$. Each space proposal is optimized for 1000 iterations in total (Eqn~\ref{eq_clock_loss}). The computational targets are set to 4.8M parameters (x-axis) and 300M FLOPs (y-axis), which are consistent with FBNet-B~\cite{fbnet}. As shown in Fig.~\ref{fig_vis_space_proposal}, with the decrease of temperature of $\tau$, the sampled architectures satisfy the computational targets more precisely. 

\iffalse
\begin{figure}[H]
	\centering
	\subfigure[$\tau=5.0$]{\includegraphics[width=0.2\linewidth]{figs/proposals/5.png}}
	\hspace{10pt}
	\subfigure[$\tau=1.0$]{\includegraphics[width=0.2\linewidth]{figs/proposals/1.png}}
	\hspace{10pt}
	\subfigure[$\tau=0.5$]{\includegraphics[width=0.2\linewidth]{figs/proposals/05.png}}
	\hspace{10pt}
	\subfigure[$\tau=0.1$]{\includegraphics[width=0.2\linewidth]{figs/proposals/01.png}}
	\caption{The distribution of 10,000 architectures sampled from optimized space proposal under different temperatures $\tau$.}
	\label{fig_vis_space_proposal}
\end{figure}
\fi

\begin{figure}
	\centering
	\subfigure[$\tau=5.0$]{\includegraphics[width=0.24\linewidth]{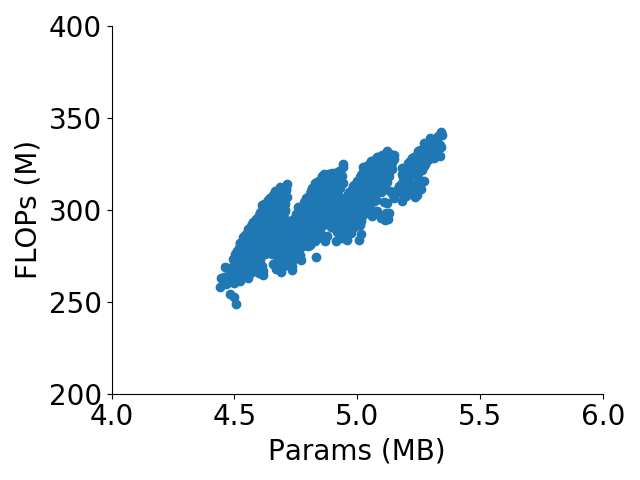}}
	\subfigure[$\tau=1.0$]{\includegraphics[width=0.24\linewidth]{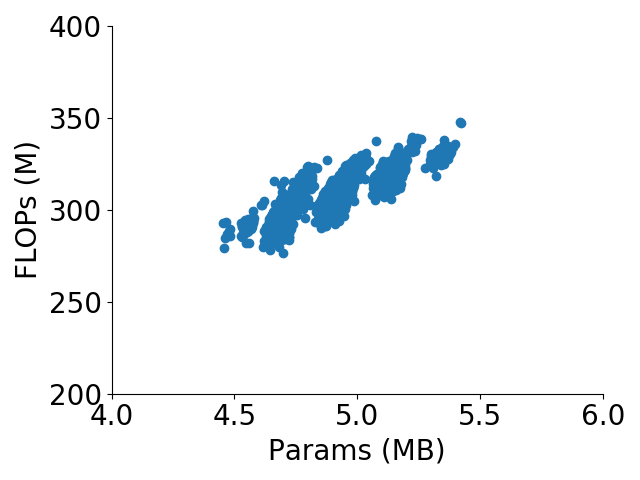}}
	\subfigure[$\tau=0.5$]{\includegraphics[width=0.24\linewidth]{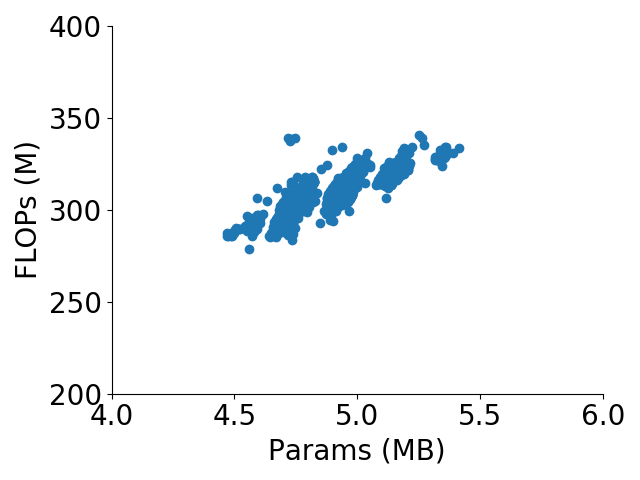}}
	\subfigure[$\tau=0.1$]{\includegraphics[width=0.24\linewidth]{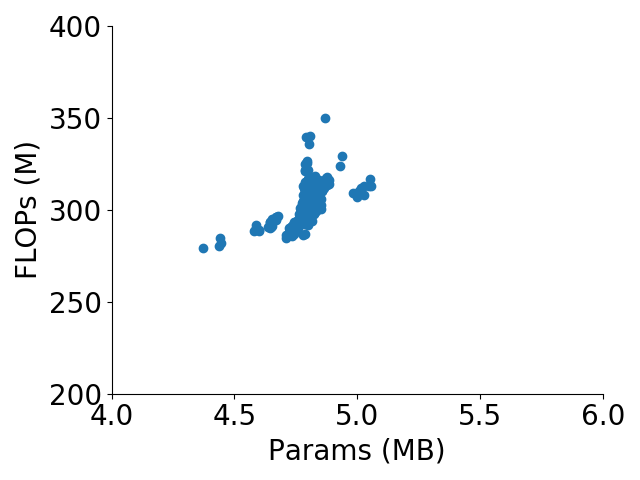}}
	\caption{The distribution of 10,000 architectures sampled from optimized space proposal under different temperatures $\tau$.}
	\label{fig_vis_space_proposal}
	\vspace{-10pt}
\end{figure}

The optimization step for constructing several space proposals takes only a few seconds, which is an efficient solution for controllable multi-objective neural architecture search. We enumerate $m= \{1, 2, 4, 8, 16\}$ and the operations with the highest probability according to $\Pi$ and $\Theta$ are selected after searching. The architectures are evaluated on the CIFAR-10 dataset to determine the appropriate space proposal number $m$ for finding superior architectures. {\color{black} In retraining, we integrate CutOut~\cite{cutout} to make networks generalize better.} As shown in Tab.~\ref{tab_appropriate_m}, $m=8$ could result in a well-performed architecture and we use $m=8$ in the following experiments to achieve a better trade-off between searching costs and performance.

\begin{table}[H]
	\caption{Comparison of image classifiers on CIFAR-10 dataset.}
	\small
	\centering
	\begin{tabular}{l|c|c|c}
		\toprule
		\multirow{2}{*}{{Model}} & Test Error & {Params} & Search Cost  \\
		& (\%)       & (M)      & (GPU days) \\
		\midrule
		%DenseNet-BC~\cite{densenet} & 3.46 & 25.6 & - &  manual \\
		%\midrule
		%PNAS~\cite{pnas} & 3.41 & 3.2 & 225 &  SMBO \\
		%AmoebaNet-A + cutout~\cite{amoebanet} & 3.12 & 3.1 & 3150 &  evolution \\
		%Hierarchical evolution~\cite{hierarchical} & 3.75 & 15.7 & 300 &  evolution \\
		%\midrule
		HourNAS~(m=1)  & 3.86 & 2.8 & 0.1  \\
		HourNAS~(m=2)  & 3.54 & 2.8 & 0.1  \\
		HourNAS~(m=4)  & 3.41 & 2.7 & 0.1  \\
		HourNAS~(m=8)  & 3.39 & 2.8 & 0.1  \\
		HourNAS~(m=16)  & 3.37 & 2.8 & 0.1  \\
		\bottomrule
	\end{tabular}
	\label{tab_appropriate_m}
	\vspace{-15pt}
\end{table}

We search for models on the ImageNet dataset, \ie, HourNAS-FBNetSS-A and HourNAS-FBNetSS-B. To fairly compare with FBNet, HourNAS-FBNetSS-A has the same model size and FLOPs as FBNet-B, and HourNAS-FBNetSS-B has the same computational requirements as FBNet-C. {\color{black} We train the networks for 350 epochs in total with a batch size of 512. Learning rate starts from 0.25 and the weight decay is 1e-5. Label smoothing and learning rate warmup strategies are also used. The activation after convolution is the ReLU function.} The training strategy is the same as FBNet~\cite{fbnet} without using bells and whistles. As shown in Tab.~\ref{tab_imagenet}, the HourNAS-FBNetSS-A and HourNAS-FBNetSS-B achieve competitive accuracies with FBNet-B and FBNet-C, and the search time is drastically reduced by two orders of magnitude. We search HourNAS-FBNetSS-A for three times using different random seeds, and the standard deviation of Top-1 accuracy is 0.1\%.

\textbf{Enlarged FBNet Search Space (EFBNetSS).} MixConv~\cite{mixconv} indicates that a larger kernel size leads to better performance. To understand the impact of search space and to further verify the effectiveness of HourNAS, we slightly enlarge the search space of FBNet. We add the blocks with kernel size $k=7$ and remove the blocks with group $g=2$. {\color{black} This modification results in a search space containing $1\times 10^{22}$ architectures, which is 10 times larger than the original one.} The multi-objectives are the same as HourNAS-FBNetSS-A and HourNAS-FBNetSS-B. We list the searched architectures in Tab.~\ref{tab_imagenet}. The HourNAS-EFBNetSS-C achieves the same Top-1 accuracy with HourNAS-FBNetSS-A and HourNAS-EFBNetSS-D surpasses HourNAS-FBNetSS-B by 0.3\% Top-1 accuracy. The larger kernel size $k=7$ ensures that the architectures are capable of perceiving the characteristics of a larger area.

\begin{table*}
	\caption{The results of FBNet-Max and EfficientNet-Max on ILSVRC2012 dataset.}
	\centering
	\small
	\begin{tabular}{c|c|c|c|c}
		\toprule
		{Model} & Params (M) & FLOPS (M) & Top-1 (\%) & Top-5 (\%) \\
		\midrule
		FBNet-Max & 5.7 & 583 & 75.7 & 92.8 \\
		\midrule
		EfficientNet-Max & 5.8 & 738 & {78.3} & {94.0} \\
		\bottomrule
	\end{tabular}
	%\end{adjustwidth}
	\label{tab_full}
	\vspace{10pt}
	
	\caption{Comparisons of searching with and without vital block priori on ILSVRC2012 dataset. The search spaces are original (upper) and enlarged (lower) FBNet search space, respectively.}
	\centering
	\small
	%\begin{adjustwidth}{-6pt}{}
	\begin{tabular}{c|c|c|c|c|c|c|c}
		\toprule
		\multirow{2}{*}{{Model}} & \multirow{2}{*}{{Type}} & Search & Search Cost & Params & FLOPS & Top-1 & Top-5 \\
		& & Dataset & (GPU days) & (M) & (M) & (\%) & (\%) \\
		\midrule
		\textbf{HourNAS-FBNetSS-A} & auto & ImageNet & \textbf{0.1} & 4.8 & 298 & \textbf{74.1} & \textbf{91.8} \\
		{HourNAS-FBNetSS-G}~(w/o vital block priori) & auto & ImageNet & {0.2} & 4.7 & 297 & 73.2 & 91.4 \\
		\midrule
		\textbf{HourNAS-EFBNetSS-D} & auto & ImageNet & \textbf{0.1} & 4.8 & 296 & \textbf{74.1} & \textbf{91.6} \\
		{HourNAS-EFBNetSS-H}~(w/o vital block priori) & auto & ImageNet & {0.2} & 4.8 & 299 & {73.5} & {91.3} \\
		\bottomrule
	\end{tabular}
	%\end{adjustwidth}
	\label{tab_vital_priori}
	\vspace{10pt}
	
	%\
	\caption{The results comparison on ILSVRC2012 dataset. }
	\centering
	\small
	%\begin{adjustwidth}{-6pt}{}
	\begin{tabular}{c|c|c|c|c|c|c|c}
		\toprule
		\multirow{2}{*}{{Model}} & \multirow{2}{*}{{Type}} & Search & Search Cost & Params & FLOPS & Top-1 & Top-5 \\
		& & Dataset & (GPU days) & (M) & (M) & (\%) & (\%) \\
		\midrule
		{HourNAS-FBNetSS-A} & auto & ImageNet & {0.1} & 4.8 & 298 & {74.1} & {91.8} \\
		{HourNAS-FBNetSS-I} & auto & ImageNet & 1.0 & 4.8 & 318 & 74.2 & 91.8 \\
		\bottomrule
	\end{tabular}
	%\end{adjustwidth}
	\label{tab_gumbel}
	\vspace{-15pt}
\end{table*}

\textbf{MnasNet  Search Space (MnasNetSS).} We further apply our proposed HourNAS to the search space of MnasNet~\cite{mnasnet}. {\color{black} The search space contains $2.5 \times 10^{23}$ architectures in total and is larger than FBNet search space.} We select MnasNet-A1 as the baseline and use its number of the parameters and FLOPs as two objectives to optimize 8 space proposals. The discovered HourNAS-MnasNetSS-E achieves a Top-1 accuracy of 75.7\% on the ILSVRC2012 dataset, which surpasses MnasNet-A1 by 0.5\%.

\textbf{EfficientNet  Search Space (EfficientNetSS).} To compare with the state-of-the-art architecture EfficientNet-B0~\cite{efficientnet}, we also use HourNAS to  search on the same search space as EfficientNet, which {\color{black} contains $4 \times 10^{18}$ architectures.} Targeting at EfficientNet-B0, we use its model size and FLOPs as two objectives to regularize space proposals and name the searched architecture as HourNAS-EfficientNetSS-F. {\color{black} Same as EfficientNet\footnote{https://github.com/tensorflow/tpu/tree/master/models/official/efficientnet}, we use the Swish~\cite{searchactivation} activation and Exponential Moving Average~(EMA) in fully training.} Note that the AutoAugment~\cite{autoaugment} is not used.  The result in Tab.~\ref{tab_imagenet} shows HourNAS-EfficientNetSS-F surpasses EfficientNet-B0 by 0.2\% Top-1 accuracy with similar number of parameters and FLOPs.

\subsection{Ablation Study}\label{sec_ablation}

If we do not restrict the computational resources of the sampled architectures in searching, the most complex block achieves the highest probability after searching for enough time. As shown in Tab.~\ref{tab_full}, we train the most complex architectures in both FBNet and EfficientNet search spaces, namely FBNet-Max and EfficientNet-Max. These two models obtain 75.7\%, and 78.3\% Top-1 accuracies, respectively. However, the computational resource requirements of these structures are relatively high. Therefore, the neural architecture search~(NAS) could be regarded as the problem of computational resource allocation given the resource constraints.

\textbf{The Impact of Vital Block Priori.} In order to investigate the impact of the vital block priori, we directly search architectures without using the vital block information. All the blocks in the SuperNet $S$ are treated equally in searching. We use the Gumbel-Max sampling and space proposal strategy to search architectures under the same predefined computational resources.

We use the previously described original and enlarged FBNet search spaces. We optimize 8 space proposals and it takes 6 hours for searching, which is twice of the counterpart that utilize the vital block priori. As shown in Tab.~\ref{tab_vital_priori}, the Top-1 accuracies of the discovered models (HourNAS-FBNetSS-G, HourNAS-EFBNetSS-H) drop by 0.9\% and 0.6\% on the ImageNet validation set, respectively. The searched vital blocks of HourNAS-FBNetSS-A uses 0.9M parameters and 130M FLOPs, and the HourNAS-FBNetSS-G uses 0.5M parameters and 55M FLOPs. The vital blocks in HourNAS-FBNetSS-G are not as expressive as HourNAS-FBNetSS-A, which results in worse performance. The results show the necessity of the vital block priori. Searching the vital blocks with higher priority is helpful in finding high-quality architectures. Therefore, we use a two-stage search method to allocate resources to vital blocks first, which can allocate resources more effectively, so as to complete the architecture search in a short time. The architectures are provided in the supplementary file, under same computational resources constraints, inclining more resources on the vital blocks gains more performance profit.

\textbf{The Impact of Gumbel-Max Sampling.} As discussed in Sec.~\ref{sec_space_proposal}, there are several design choices for the sampling methods. To find out the impact of the Gumbel-Max sampling method, here we instead use the Gumbel softmax~\cite{fbnet} to optimize architecture parameters and network parameters. The search space and target resource constraints are the same as HourNAS-FBNetSS-A. The search process takes around 1 GPU day and the finalized architecture is denoted as HourNAS-FBNetSS-I. As shown in Tab.~\ref{tab_gumbel}, HourNAS-FBNetSS-I outperforms HourNAS-FBNetSS-A by 0.1\% Top-1 accuracy with much less searching cost, which demonstrate that Gumbel-Max is an efficient strategy for optimizing the SuperNet with almost no less of accuracy.

\section{Conclusions}\label{sec:con}

This paper investigates an efficient algorithm to search deep neural architectures on the large-scale dataset (\emph{i.e.}, ImageNet) directly. To reduce the complexity of the huge search space, we present an Hourglass-based search framework, namely HourNAS. The entire search space is divided into ``vital'' and ``non-vital'' parts accordingly. By gradually search the components in each part, the search cost can be reduced significantly. Since the ``vital'' parts are more important for the performance of the obtained neural network, the optimization on this part can ensure accuracy. By exploiting the proposed approach, we can directly search architectures on the ImageNet dataset that achieves a 77.0\% Top-1 accuracy using only 3 hours (\emph{i.e.}, about 0.1 GPU days), which outperforms the state-of-the-art methods in both terms of search speed and accuracy.

\clearpage
\newpage

{
\bibliographystyle{ieee_fullname}
\bibliography{nas}

\begin{thebibliography}{10}\itemsep=-1pt

\bibitem{tunas}
Gabriel Bender, Hanxiao Liu, Bo Chen, Grace Chu, Shuyang Cheng, Pieter-Jan
  Kindermans, and Quoc~V Le.
\newblock Can weight sharing outperform random architecture search? an
  investigation with tunas.
\newblock {\em CVPR}.

\bibitem{onceforall}
Han {Cai}, Chuang {Gan}, Tianzhe {Wang}, Zhekai {Zhang}, and Song {Han}.
\newblock Once for all: Train one network and specialize it for efficient
  deployment.
\newblock {\em ICLR}, 2020.

\bibitem{proxylessnas}
Han Cai, Ligeng Zhu, and Song Han.
\newblock Proxylessnas: Direct neural architecture search on target task and
  hardware.
\newblock {\em ICLR}, 2019.

\bibitem{data}
Jianlong {Chang}, xinbang {zhang}, Yiwen {Guo}, Gaofeng {Meng}, Shiming
  {Xiang}, and Chunhong {Pan}.
\newblock Data: Differentiable architecture approximation.
\newblock {\em NeurIPS}, 2019.

\bibitem{pdarts}
Xin Chen, Lingxi Xie, Jun Wu, and Qi Tian.
\newblock Progressive differentiable architecture search: Bridging the depth
  gap between search and evaluation.
\newblock {\em ICCV}, 2019.

\bibitem{multicolumndnn}
Dan Ciregan, Ueli Meier, and J{\"u}rgen Schmidhuber.
\newblock Multi-column deep neural networks for image classification.
\newblock {\em CVPR}.

\bibitem{autoaugment}
Ekin~D Cubuk, Barret Zoph, Dandelion Mane, Vijay Vasudevan, and Quoc~V Le.
\newblock Autoaugment: Learning augmentation strategies from data.
\newblock {\em CVPR}, 2019.

\bibitem{fpnas}
Jiequan Cui, Pengguang Chen, Ruiyu Li, Shu Liu, Xiaoyong Shen, and Jiaya Jia.
\newblock Fast and practical neural architecture search.
\newblock {\em ICCV}, 2019.

\bibitem{cutout}
Terrance DeVries and Graham~W Taylor.
\newblock Improved regularization of convolutional neural networks with cutout.
\newblock {\em arXiv}, 2017.

\bibitem{globalsparsemomentumsgd}
Xiaohan {Ding}, guiguang {ding}, Xiangxin {Zhou}, Yuchen {Guo}, Jungong {Han},
  and Ji {Liu}.
\newblock Global sparse momentum sgd for pruning very deep neural networks.
\newblock {\em NeurIPS}, 2019.

\bibitem{setnas}
Xuanyi Dong and Yi Yang.
\newblock One-shot neural architecture search via self-evaluated template
  network.
\newblock {\em ICCV}, 2019.

\bibitem{gdas}
Xuanyi {Dong} and Yi {Yang}.
\newblock Searching for a robust neural architecture in four gpu hours.
\newblock {\em CVPR}, 2019.

\bibitem{nasbench201}
Xuanyi Dong and Yi Yang.
\newblock Nas-bench-102: Extending the scope of reproducible neural
  architecture search.
\newblock {\em ICLR}, 2020.

\bibitem{lemonade}
Thomas {Elsken}, Jan {Metzen}, and Frank {Hutter}.
\newblock Efficient multi-objective neural architecture search via lamarckian
  evolution.
\newblock {\em ICLR}, 2019.

\bibitem{frcnn}
Ross Girshick.
\newblock Fast r-cnn.
\newblock {\em ICCV}, 2015.

\bibitem{autogan}
Xinyu Gong, Shiyu Chang, Yifan Jiang, and Zhangyang Wang.
\newblock Autogan: Neural architecture search for generative adversarial
  networks.
\newblock {\em ICCV}, 2019.

\bibitem{sposnas}
Zichao {Guo}, Xiangyu {Zhang}, Haoyuan {Mu}, Wen {Heng}, Zechun {Liu}, Yichen
  {Wei}, and Jian {Sun}.
\newblock Single path one-shot neural architecture search with uniform
  sampling.
\newblock {\em arXiv}, 2019.

\bibitem{maskrcnn}
Kaiming He, Georgia Gkioxari, Piotr Doll{\'a}r, and Ross Girshick.
\newblock Mask r-cnn.
\newblock {\em ICCV}, 2017.

\bibitem{resnet}
Kaiming He, Xiangyu Zhang, Shaoqing Ren, and Jian Sun.
\newblock Deep residual learning for image recognition.
\newblock {\em CVPR}, 2016.

\bibitem{amc}
Yihui {He}, Ji {Lin}, Zhijian {Liu}, Hanrui {Wang}, Li-Jia {Li}, and Song
  {Han}.
\newblock Amc: Automl for model compression and acceleration on mobile devices.
\newblock {\em ECCV}, 2018.

\bibitem{mobilenet}
Andrew~G. {Howard}, Menglong {Zhu}, Bo {Chen}, Dmitry {Kalenichenko}, Weijun
  {Wang}, Tobias {Weyand}, Marco {Andreetto}, and Hartwig {Adam}.
\newblock Mobilenets: Efficient convolutional neural networks for mobile vision
  applications.
\newblock {\em arXiv}, 2017.

\bibitem{densenet}
Gao Huang, Zhuang Liu, Laurens Van Der~Maaten, and Kilian~Q Weinberger.
\newblock Densely connected convolutional networks.
\newblock {\em CVPR}, 2017.

\bibitem{stochasticdepth}
Gao Huang, Yu Sun, Zhuang Liu, Daniel Sedra, and Kilian~Q Weinberger.
\newblock Deep networks with stochastic depth.
\newblock {\em ECCV}.

\bibitem{adam}
Diederik~P. {Kingma} and Jimmy~Lei {Ba}.
\newblock Adam: A method for stochastic optimization.
\newblock {\em ICLR}, 2015.

\bibitem{cifar}
Alex Krizhevsky and Geoffrey Hinton.
\newblock Learning multiple layers of features from tiny images.
\newblock Technical report, 2009.

\bibitem{alexnet}
Alex Krizhevsky, Ilya Sutskever, and Geoffrey~E Hinton.
\newblock Imagenet classification with deep convolutional neural networks.
\newblock {\em NeurIPS}, 2012.

\bibitem{li2020random}
Liam Li and Ameet Talwalkar.
\newblock Random search and reproducibility for neural architecture search.
\newblock {\em UAI}.

\bibitem{autodeeplab}
Chenxi Liu, Liang-Chieh Chen, Florian Schroff, Hartwig Adam, Wei Hua, Alan~L
  Yuille, and Li Fei-Fei.
\newblock Auto-deeplab: Hierarchical neural architecture search for semantic
  image segmentation.
\newblock {\em CVPR}, 2019.

\bibitem{pnas}
Chenxi Liu, Barret Zoph, Maxim Neumann, Jonathon Shlens, Wei Hua, Li-Jia Li, Li
  Fei-Fei, Alan Yuille, Jonathan Huang, and Kevin Murphy.
\newblock Progressive neural architecture search.
\newblock {\em ECCV}, 2018.

\bibitem{darts}
Hanxiao {Liu}, Karen {Simonyan}, and Yiming {Yang}.
\newblock Darts: Differentiable architecture search.
\newblock {\em ICLR}, 2019.

\bibitem{ssd}
Wei Liu, Dragomir Anguelov, Dumitru Erhan, Christian Szegedy, Scott Reed,
  Cheng-Yang Fu, and Alexander~C Berg.
\newblock Ssd: Single shot multibox detector.
\newblock {\em ECCV}, 2016.

\bibitem{nsganet}
Zhichao {Lu}, Ian {Whalen}, Vishnu {Boddeti}, Yashesh~D. {Dhebar}, Kalyanmoy
  {Deb}, Erik~D. {Goodman}, and Wolfgang {Banzhaf}.
\newblock Nsga-net: A multi-objective genetic algorithm for neural architecture
  search.
\newblock {\em GECC}, 2018.

\bibitem{shufflenetv2}
Ningning Ma, Xiangyu Zhang, Hai-Tao Zheng, and Jian Sun.
\newblock Shufflenet v2: Practical guidelines for efficient cnn architecture
  design.
\newblock {\em ECCV}, 2018.

\bibitem{astarsampling}
Chris~J Maddison, Daniel Tarlow, and Tom Minka.
\newblock A* sampling.
\newblock {\em NeurIPS}, 2014.

\bibitem{enas}
Hieu Pham, Melody~Y Guan, Barret Zoph, Quoc~V Le, and Jeff Dean.
\newblock Efficient neural architecture search via parameter sharing.
\newblock {\em ICML}, 2018.

\bibitem{autoreid}
Ruijie Quan, Xuanyi Dong, Yu Wu, Linchao Zhu, and Yi Yang.
\newblock Auto-reid: Searching for a part-aware convnet for person
  re-identification.
\newblock {\em ICCV}, 2019.

\bibitem{searchactivation}
Prajit Ramachandran, Barret Zoph, and Quoc~V Le.
\newblock Searching for activation functions.
\newblock {\em arXiv}, 2017.

\bibitem{amoebanet}
Esteban Real, Alok Aggarwal, Yanping Huang, and Quoc~V Le.
\newblock Regularized evolution for image classifier architecture search.
\newblock {\em AAAI}, 2019.

\bibitem{mobilenetv2}
Mark~B. {Sandler}, Andrew~G. {Howard}, Menglong {Zhu}, Andrey {Zhmoginov}, and
  Liang-Chieh {Chen}.
\newblock Mobilenetv2: Inverted residuals and linear bottlenecks.
\newblock {\em CVPR}, 2018.

\bibitem{metaarchitecturesearch}
Albert {Shaw}, Wei {Wei}, Weiyang {Liu}, Le {Song}, and Bo {Dai}.
\newblock Meta architecture search.
\newblock {\em NeurIPS}, 2019.

\bibitem{vgg}
Karen Simonyan and Andrew Zisserman.
\newblock Very deep convolutional networks for large-scale image recognition.
\newblock {\em ICLR}, 2015.

\bibitem{mnasnet}
Mingxing {Tan}, Bo {Chen}, Ruoming {Pang}, Vijay {Vasudevan}, Mark {Sandler},
  Andrew {Howard}, and Quoc~V. {Le}.
\newblock Mnasnet: Platform-aware neural architecture search for mobile.
\newblock {\em CVPR}, 2018.

\bibitem{efficientnet}
Mingxing {Tan} and Quoc~V. {Le}.
\newblock Efficientnet: Rethinking model scaling for convolutional neural
  networks.
\newblock {\em ICML}, 2019.

\bibitem{mixconv}
Mingxing {Tan} and Quoc~V. {Le}.
\newblock Mixconv: Mixed depthwise convolutional kernels.
\newblock {\em BMVC}, 2019.

\bibitem{resnetensembles}
Andreas {Veit}, Michael {Wilber}, and Serge {Belongie}.
\newblock Residual networks behave like ensembles of relatively shallow
  networks.
\newblock {\em NeurIPS}, 2016.

\bibitem{fbnet}
Bichen {Wu}, Xiaoliang {Dai}, Peizhao {Zhang}, Yanghan {Wang}, Fei {Sun},
  Yiming {Wu}, Yuandong {Tian}, Peter {Vajda}, Yangqing {Jia}, and Kurt
  {Keutzer}.
\newblock Fbnet: Hardware-aware efficient convnet design via differentiable
  neural architecture search.
\newblock {\em CVPR}, 2019.

\bibitem{snas}
Sirui Xie, Hehui Zheng, Chunxiao Liu, and Liang Lin.
\newblock Snas: stochastic neural architecture search.
\newblock {\em ICLR}, 2019.

\bibitem{resourceconstrainednas}
Yunyang Xiong, Ronak Mehta, and Vikas Singh.
\newblock Resource constrained neural network architecture search.
\newblock {\em ICCV}, 2019.

\bibitem{yang2019evaluation}
Antoine Yang, Pedro~M Esperan{\c{c}}a, and Fabio~M Carlucci.
\newblock Nas evaluation is frustratingly hard.
\newblock {\em ICLR}, 2020.

\bibitem{cars}
Zhaohui Yang, Yunhe Wang, Xinghao Chen, Boxin Shi, Chao Xu, Chunjing Xu, Qi
  Tian, and Chang Xu.
\newblock Cars: Continuous evolution for efficient neural architecture search.
\newblock {\em CVPR}, 2020.

\bibitem{nasbench101}
Chris Ying, Aaron Klein, Eric Christiansen, Esteban Real, Kevin Murphy, and
  Frank Hutter.
\newblock Nas-bench-101: Towards reproducible neural architecture search.
\newblock {\em ICML}.

\bibitem{yu2019evaluating}
Kaicheng Yu, Christian Sciuto, Martin Jaggi, Claudiu Musat, and Mathieu
  Salzmann.
\newblock Evaluating the search phase of neural architecture search.
\newblock {\em ICLR}, 2019.

\bibitem{undersandingrobustfyingnas}
Arber {Zela}, Thomas {Elsken}, Tonmoy {Saikia}, Yassine {Marrakchi}, Thomas
  {Brox}, and Frank {Hutter}.
\newblock Understanding and robustifying differentiable architecture search.
\newblock {\em ICLR}, 2020.

\bibitem{arealllayerscreatedequal}
Chiyuan {Zhang}, Samy {Bengio}, and Yoram {Singer}.
\newblock Are all layers created equal.
\newblock {\em ICMLW}, 2019.

\bibitem{mdenas}
Xiawu Zheng, Rongrong Ji, Lang Tang, Baochang Zhang, Jianzhuang Liu, and Qi
  Tian.
\newblock Multinomial distribution learning for effective neural architecture
  search.
\newblock {\em ICCV}, 2019.

\bibitem{zoph_rl_iclr}
Barret Zoph and Quoc~V Le.
\newblock Neural architecture search with reinforcement learning.
\newblock {\em ICLR}, 2017.

\bibitem{nasnet}
Barret Zoph, Vijay Vasudevan, Jonathon Shlens, and Quoc~V Le.
\newblock Learning transferable architectures for scalable image recognition.
\newblock {\em CVPR}, 2018.

\end{thebibliography}
}

\end{document}